\documentclass{article}

\usepackage[numbers]{natbib}
\usepackage{tcolorbox}
\usepackage[dblblindworkshop, final]{neurips_2026}
\usepackage{amsmath}

\usepackage[dblblindworkshop, final]{neurips_2026}

\workshoptitle{Resource-Aware Agentic AI}

\usepackage[utf8]{inputenc} 
\usepackage[T1]{fontenc}    
\usepackage{hyperref}       
\usepackage{url}            
\usepackage{booktabs}       
\usepackage{amsfonts}       
\usepackage{nicefrac}       
\usepackage{microtype}      
\usepackage{xcolor}  
\usepackage{hyperref}
\usepackage{url}
\usepackage{multirow}
\usepackage[normalem]{ulem}
\usepackage{graphicx}
\usepackage{wrapfig}
\usepackage{placeins}
\usepackage{tabularx}
\usepackage{booktabs}

\title{Blind Spots: Where Small Language Models Fall Behind Frontier Models on Terminal Tasks}

\newcommand{\aspace}{\hspace{2.05em}}
 \newcommand{\ucla}{$^{\heartsuit}$}
\newcommand{\msft}{$^{\clubsuit}$}

\author{%
    Ashima Suvarna\ucla\thanks{Work done during Ashima Suvarna’s internship at Microsoft Research. \msft Equal mentorship.}\aspace
    Ruta Desai\msft\aspace
    Ahmed Elgohary\msft\\
    \msft Microsoft Research \aspace \ucla University of California, Los Angeles \\ 
    \texttt{asuvarna31@g.ucla.edu}, \texttt{\{rutadesai,ahmedghoneim\}@microsoft.com}\\
}

\newcommand{\tmaxnine}{Tmax-9B}

\newcommand{\qwennine}{Qwen3.5-9B}

\newcommand{\glm}{GLM-5.2}

\newcommand{\minisa}{mini-swe-agent}
\newcommand{\vanillux}{vanillux}
\newcommand{\terminus}{terminus-2}
\newcommand{\openhands}{OpenHands}

\newcommand{\tblite}{\textsc{ot-tblite}}
\newcommand{\tbtwo}{\textsc{tb2.1}}

\renewcommand{\vanillux}{vanillux2agent} 
\renewcommand{\tblite}{\mbox{\textsc{tblite}}} 
\renewcommand{\qwennine}{\mbox{Qwen3.5-9B}}
\renewcommand{\tmaxnine}{\mbox{Tmax-9B}}
\renewcommand{\glm}{\mbox{GLM-5.2}}

\begin{document}

\maketitle

\begin{abstract}

Small language models (SLMs) offer an attractive alternative to frontier models (FMs) because they can be deployed with lower cost and latency. This makes SLMs especially appealing for terminal tasks on edge devices, where responsiveness, local execution, and limited connectivity are important considerations. However, they continue to lag behind FMs on terminal tasks,
as measured by standard task success rates. Yet these rates conflate several capabilities that may contribute to this gap. In this work, we characterize the \mbox{SLM--FM} gap along three axes: \emph{solvability}, whether additional test-time compute allows an SLM to reach a correct solution; \emph{verification}, whether it can identify a successful solution among candidate trajectories; and \emph{confidence}, whether it can accurately anticipate its probability of success. 
Across two standard terminal benchmarks,
we find that repeated sampling narrows, but does not eliminate, the gap in solvability. In contrast, a persistent gap remains in verification. Given the same candidate trajectory pool, FMs recover substantially more of the gap between random and oracle selection than SLMs. SLMs are also more overconfident in their prospective success estimates and exhibit poorer calibration. These results suggest that the \mbox{SLM--FM} gap is multidimensional, spanning not only the ability to generate successful solutions, but also the ability to recognize successful solutions and anticipate success. We hope this decomposition can guide future work on developing stronger and smaller terminal agents.

\end{abstract}

\begin{figure}[h]
    \centering
    \includegraphics[width=0.8\linewidth]{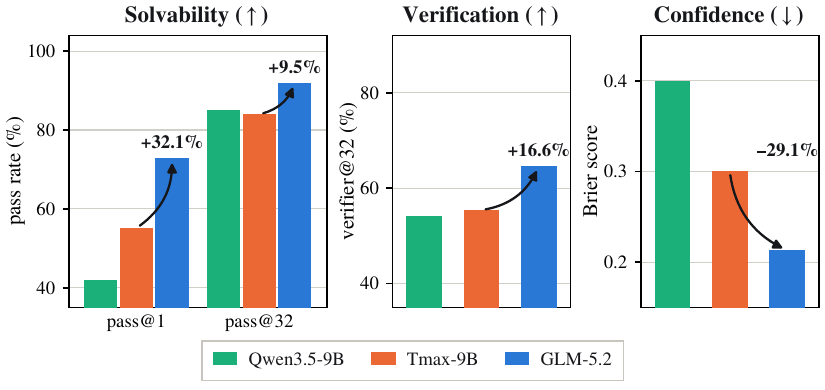}
    \caption{ \textbf{The \mbox{SLM--FM} gap along three axes on \tblite{}.} Solvability improves substantially with repeated sampling, narrowing the performance gap; outcome verification remains substantially weaker for SLMs; and SLMs are less calibrated in their prospective estimates of task success. Annotated arrows indicate the relative gap between \glm{} and \tmaxnine{}.}

    \label{fig:summary}
\end{figure}

\section{Introduction}

Terminal agents solve long-horizon and complex computer-use tasks---configuring systems, fixing repositories, processing data, managing
services---by executing shell commands, writing and running code, and
interacting with the environment in a multi-step fashion. In recent years, terminal agents have shown remarkable progress~\cite{claudecode2025, codex2025, devin2024,glm2026glm5, anthropic2026opus5}. While frontier models (FMs) continue to lead on the terminal benchmark leaderboards~\cite{tbench_leaderboard}, significant progress has been made in training highly capable small language models (SLMs)~\cite{ivison2026tmax, raoof2026openthoughts, pi2026data, shrivastava2026echo, peng2026litecoderterminal}. SLMs are attractive as terminal agents because they can be deployed locally at low cost and low latency. Nevertheless, a wide gap remains. On Terminal-Bench 2.1~(\tbtwo{}), the best frontier
systems on the leaderboard reach 87.4\% task success while the strongest small models
reach 28.8\%\cite{tbench_leaderboard, ivison2026tmax}. 

This gap in task success rates conflates distinct capabilities and says little about why trials fail. Our work aims to understand what constitutes this gap. We study failed trajectories from both model tiers using a common failure taxonomy~\cite{merrill2026terminal}. Our analysis reveals that SLMs and FMs exhibit broadly the same major failure categories---step repetition, weak verification, and disobeying the specification---but SLMs experience these failures more frequently rather than in qualitatively different ways (\S\ref{sec:failures}). This suggests that the \mbox{SLM--FM} gap reflects differences in capabilities that cut across these shared failure modes. Motivated by these patterns, we examine three complementary capabilities: \emph{solvability}, the ability to reach a successful solution given sufficient attempts; \emph{verification}, the ability to distinguish successful from unsuccessful solutions; and \emph{confidence}, the ability to assess one's likelihood of success before acting. We share qualitative examples to demonstrate each axis in Table~\ref{tab:qualitative}. Distinguishing these capabilities is important both for understanding the source of the performance gap and for identifying effective techniques to improve SLMs.

Our central finding is that repeated sampling narrows the \mbox{SLM--FM} gap in solvability but not in verification or confidence: with more attempts, small models increasingly find correct solutions but still struggle to recognize them or anticipate when they will succeed (Figure~\ref{fig:summary}). On OpenThoughts-TBLite (\tblite{})~\cite{raoof2026openthoughts}, the absolute gap between \glm{} and \tmaxnine{} shrinks from 17.7 percentage points at pass@1 (72.8 for \glm{} vs.\ 55.1 for \tmaxnine{}) to 8.0 points at pass@32 (92.0 vs.\ 84.0, respectively), and the relative gap shrinks from 32.1\% to 9.5\% (Figure~\ref{fig:summary}).
In contrast, the
verification gap persists even when successful trajectories are available.
Given the same candidate pool of 32 trajectories, \glm{} selects a successful trajectory
more often than \tmaxnine{} at $k{=}32$, an absolute gap of 9.2 points (64.6 vs.\ 55.4) and a relative gap of 16.6\%. A similar pattern emerges in confidence. \glm{}'s prospective success estimates are better calibrated,
with a Brier score 29.1\% lower than that of \tmaxnine{} in relative terms.
Together, these results suggest that additional test-time compute helps
SLMs reach correct solutions, but does not close the gap in recognizing
those solutions or in anticipating their own success.

\begin{table}
\centering
\footnotesize
\setlength{\tabcolsep}{4pt}
\renewcommand{\arraystretch}{1.15}
\begin{tabular}{@{}p{2.1cm}p{1.5cm}p{1.9cm}p{4.4cm}p{2.5cm}@{}}
\toprule
Task & Model & Solvability & Verification & Confidence \\
\midrule
\texttt{mlflow-} \texttt{register}
  & Qwen3.5-9B & solves $1/32$ & scores $31/32$ attempts as success; its chosen solution passes $3\%$ of the time & predicts $18\%$ \\
  & GLM-5.2    & solves $29/32$ & its chosen solution passes 100\% of the time
  & predicts $96\%$ \\
\midrule
\addlinespace[3pt]
\texttt{token-auth-} \texttt{websocket}
  & Qwen3.5-9B & solves $0/32$ & --- & predicts $88\%$ \\
  & GLM-5.2    & solves $26/32$ & --- & predicts $88\%$ \\
\bottomrule
\end{tabular}
\caption{\textbf{Qualitative examples of the capability gap between \qwennine{} and \glm{} on \tbtwo{} tasks.} Solvability counts each model's own successful attempts; Verification uses the same 32-attempt \qwennine{} candidate pool for both models (\S\ref{sec:verify}), and --- marks tasks whose pool contains no successful attempt; Confidence is the model's prospective success estimate.}

\label{tab:qualitative}
\end{table}

\section{Related Work}

\textbf{Terminal agents.} Recent
research on improving terminal agents largely focuses on data scale and
environment fidelity: Tmax~\citep{ivison2026tmax},
OpenThoughts-Agent~\citep{raoof2026openthoughts}, and
Nemotron-Terminal~\citep{pi2026data} construct large-scale terminal
task collections, while TermiGen~\citep{zhu2026termigen} and
Terminal-Lego~\citep{yang2026terminallego} argue that
environment-grounded trajectories matter more than trajectory count. A
parallel line refines the training objective:
LiteCoder-Terminal~\citep{peng2026litecoderterminal} studies data and
training strategies for terminal reasoning, and
ECHO~\citep{shrivastava2026echo} applies an auxiliary cross-entropy
objective to environment-observation tokens to improve RL. Evaluation
has followed: Terminal-Bench~\citep{merrill2026terminal} provides hard,
realistic command-line tasks with a software-engineering emphasis, and
\tblite{}~\citep{raoof2026openthoughts} targets a broader domain mix
tractable for smaller models; we use both these benchmarks for our analysis. Existing work reports
performance as a single success rate on these benchmarks, which is what we decompose. A
single rate cannot distinguish a model that cannot reach a solution, one
that reaches it but cannot recognize it, and one that cannot tell in advance which it will be.

\textbf{Test-time compute and repeated sampling.}
Drawing multiple attempts and selecting among them is a standard way to convert inference compute
into accuracy~\citep{chen2021codex, brown2024monkeys}. Recent work characterizes how best to
allocate that budget across sampling, revision, and search~\citep{snell2024scalingllmtesttimecompute}.
\citet{brown2024monkeys} show that coverage grows steeply with sample count for weaker models in
particular, and that the practical bottleneck shifts from generation to selection once a correct
sample is present in the pool. We take this observation as our starting point rather than our
result: we measure how much of the SLM--FM gap survives repeated sampling on terminal tasks, and
then ask whether the selection bottleneck falls differently on the two tiers.

\textbf{Verification and selection.}
Best-of-$k$ selection with trained or prompted verifiers is well established in single-turn
reasoning~\citep{cobbe2021training, lightman2023letsverifystepstep}, as is the use of models as judges of their own
or others' outputs~\citep{zheng2023judging}. Recent work also suggests that models often cannot reliably
verify or correct their own reasoning without external
signal~\citep{huang2023large, stechly2023gpt}. These results concern short answers produced in
one shot. In the terminal setting the object being judged is a multi-turn trajectory that already
contains environment feedback---exit codes, stderr, test output---so verification is in principle
better supported than in the single-turn case. We measure whether SLMs exploit that evidence, holding
the candidate pool fixed across verifier tiers so that verification skill is separated from the
quality of what is being verified.

\textbf{Confidence and calibration.}
Language models carry some signal about their own correctness~\citep{kadavath2022language}, but verbalized confidence is typically overstated relative to realized accuracy, with the size of the
gap varying by scale and post-training~\citep{lin2022teaching, tian-etal-2023-just}. Most of this work
elicits confidence \emph{after} an answer is produced. For an agent deciding how many turns to
spend, how much to think, or whether to escalate, the useful quantity is prospective: an estimate
formed before acting~\citep{barkan2025largelanguagemodelsknow}. We elicit confidence in this
prospective form in our experiments.

\section{Preliminaries}
\label{sec:prelim}

\textbf{Small and frontier models.}
Following the model tier classification of \citet{belcak2025smalllanguagemodelsfuture}, we refer to a language model as a \emph{small language model (SLM)} if it can be deployed on a consumer device with low inference latency for a single user. A \emph{frontier model (FM)} is among the most capable language models available, with state-of-the-art or near state-of-the-art performance across multiple benchmarks. This distinction is practical rather than architectural and reflects the deployment setting considered in this work. As of 2026, models with fewer than 10B parameters could be considered SLMs under our classification. We share the model sizes of all models in our work in Appendix Table~\ref{tab:models}.

\textbf{Repeated sampling and pass@$k$.}
A prominent approach for scaling test-time compute is repeated sampling:
drawing multiple independent attempts from the model and selecting one as
the solution~\citep{chen2021codex, brown2024monkeys}. In our setting each
attempt runs in a fresh environment instance and is graded by the task's
own verifier, so attempts are independent and selection requires no
external verifier or reward model. We collect $n$ attempts per task and
estimate coverage with the unbiased estimator of \citet{chen2021codex}
and report pass@$k$ averaged over tasks. The pass@1--pass@$k$ gap
measures how much of a model's deficit consists of tasks it can solve but
does not solve reliably, as opposed to tasks outside its reach at budget~$n$.

\textbf{Outcome verification.}
We use \emph{verification} to refer to actions taken to establish whether a solution is correct. We distinguish \emph{process verification}, which occurs during a trajectory through actions such as running tests or inspecting intermediate outputs, from \emph{outcome verification}, which occurs after generation and judges whether a completed trajectory succeeded. Our main verification experiment studies outcome verification: selecting a successful trajectory from a pool of completed candidates.

\textbf{Confidence.}
We define \emph{prospective confidence} as a model's estimated probability of successfully completing a task before taking any action. Unlike verification, which assesses completed solutions, confidence measures whether a model can anticipate its own success. We elicit confidence estimates from a model as a verbalized probability given only the task description, and evaluate calibration via overconfidence (the predicted–true gap) and Brier score~\cite{barkan2025largelanguagemodelsknow, lin2022teaching, tian-etal-2023-just}.

\textbf{Terminal benchmarks.} We evaluate terminal-agent capabilities on two benchmarks: \tblite{}~\cite{raoof2026openthoughts} and \tbtwo{}~\cite{merrill2026terminal}. \tblite{} comprises 100 tasks (40 easy, 26 medium, 26 hard, and 8 extremely hard based on Claude Haiku 4.5) spanning domains such as software engineering and machine learning. \tbtwo{} comprises 89 tasks (4 easy, 55 medium, and 30 hard) spanning diverse domains, designed to evaluate capable agents on realistic, complex workflows. In our experiments, \tbtwo{} is more challenging for all evaluated models, particularly SLMs (Appendix~\ref{app:difficulty}, Figure~\ref{fig:difficulty}): \tmaxnine{} achieves a pass@1 of 55.1\% on \tblite{} versus 31.5\% on \tbtwo{}, while \qwennine{} drops from 42.0\% to 21.6\%. 

\section{Failure Modes\label{sec:failures}}
\begin{figure}
    \centering
    \includegraphics[width=0.8\linewidth]{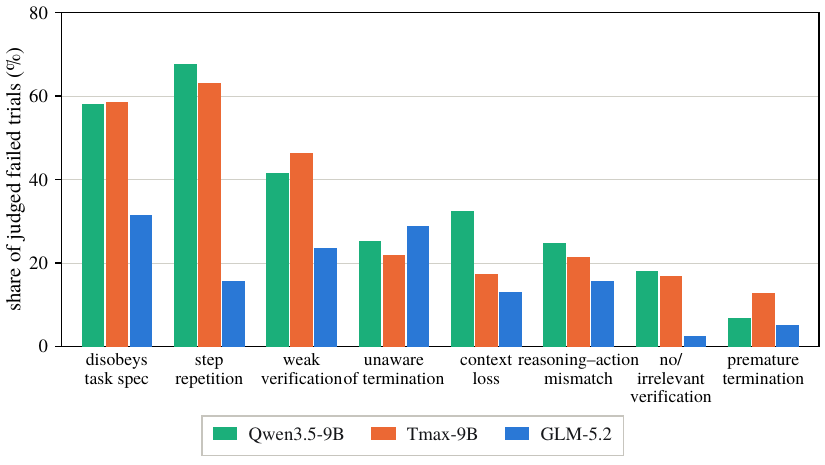}
    \caption{\textbf{SLMs and \glm{} share the same dominant failure modes.}
Share of judged failed trajectories exhibiting each failure mode on
\tblite{} under vanillux2agent.
Step repetition, disobeying the task specification, and weak
verification are among the most frequent modes for every model;
both SLMs exhibit each at higher rates than GLM-5.2. We follow the taxonomy from \citep{merrill2026terminal}.}
    \label{fig:failure_vanillux}
\end{figure}

Before decomposing the gap, we ask whether SLMs fail in qualitatively different ways than FMs. We categorize failed trajectories based on the
failure-mode taxonomy of \citet{merrill2026terminal} (defined in Appendix Table~\ref{tab:fail_modes}), using an
LLM judge (GPT-5, high reasoning). Figure~\ref{fig:failure_vanillux} shows the
share of failed trajectories exhibiting each mode under the vanillux2agent harness (expanded to other harnesses in Appendix~\ref{app:failure}). 

\textbf{Results.} We find that the dominant failure modes are shared across all models: step
repetition, disobeying the task specification, and weak verification are among the
most frequent failures for every model. SLMs exhibit all three at
substantially higher prevalence than GLM-5.2, while unawareness of termination occurs at
comparable rates across tiers. This suggests that the gap between SLMs and FMs stems from general capabilities that every terminal agent relies on, such as finding a solution, verifying it before submitting, and estimating its chances of success. These motivate the three axes we measure next.

\section{The \mbox{SLM--FM} Gap}
In this section, we decompose the \mbox{SLM--FM} gap along three axes: \emph{solvability}---whether a model can reach a correct solution with repeated sampling (\S\ref{sec:passk}); \emph{verification}---whether it can identify a correct solution from a fixed candidate pool (\S\ref{sec:verify}); and \emph{confidence}---whether it can anticipate its probability of success before acting (\S\ref{sec:confidence}). Across both \tblite{} and \tbtwo{}, we find that \textbf{additional sampling narrows the gap in solvability---substantially on \tblite{} and less so on the harder \tbtwo{}---while substantial gaps remain in verification and confidence.}

We study terminal agents built from two model tiers: SLMs and FMs. We evaluate two SLMs, \qwennine{} and \tmaxnine{}, and one FM,
\glm{}, across \tblite{} and \tbtwo{}. We consider \tmaxnine{} and its base \qwennine{} since they are among the strongest sub-10B models on terminal tasks~\cite{ivison2026tmax}. Additionally, we consider \glm{} as our FM tier since it is one of the strongest open-weights models on \tbtwo{}\footnote{Open weights allow us to
serve all models under identical conditions and harnesses.}. We use vanillux2agent~\cite{ivison2026tmax} for all main experiments. We provide a broader analysis of the effect of harness choice on model performance across four popular harnesses in Appendix~\ref{sec:harness-reward}.

\subsection{Solvability}
\label{sec:passk}

\begin{figure*}[h]
    \centering
    \includegraphics[width=0.8\linewidth]{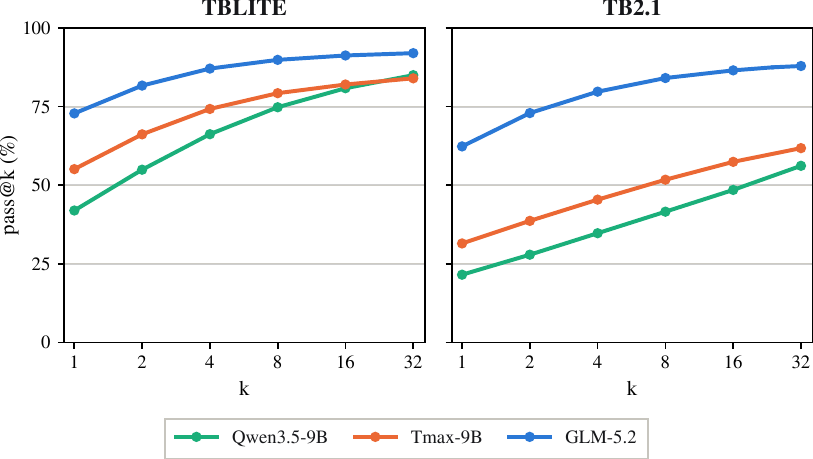}
    \caption{\textbf{Repeated sampling narrows the \mbox{SLM--FM} gap in solvability.} SLMs improve sharply with increasing $k$, while \glm{} gains less from additional attempts; the gap is reduced but remains at the largest $k$, particularly on \tbtwo{}.}
    \label{fig:passk}
\end{figure*}

Per-trial pass rates conflate tasks a model cannot solve with tasks it solves unreliably; we separate the two by scaling the attempt budget. 
We draw $n{=}32$ independent attempts per task for each model on \tblite{} and
\tbtwo{};
each attempt runs in a fresh task container and is graded by the
task's verifier. We report pass@$k$ curves for $k \leq 32$ in Figure~\ref{fig:passk}.

\textbf{Results.}
As shown in Figures~\ref{fig:summary} and~\ref{fig:passk}, repeated sampling
substantially narrows the \mbox{SLM--FM} gap. On \tblite{}, the absolute gap
between \glm{} and \tmaxnine{} is 17.7 percentage points at pass@1 and 8.0 points at
pass@32; the relative gap falls from 32.1\% to 9.5\%. In absolute terms, SLMs gain more from additional attempts (\tmaxnine{}: +28.9 points;
\qwennine{}: +43.0 points) than \glm{} (+19.2 points). At $k{=}32$, \tmaxnine{} exceeds \glm{}'s single-attempt success
on \tblite{} (84.0\% vs.\ 72.8\%) and matches it on \tbtwo{}
(61.8\% vs.\ 62.3\%). Thus, repeated sampling narrows the \mbox{SLM--FM} gap by increasing coverage: \textbf{small models can solve terminal tasks when given more attempts.}

\subsection{Verification}
\label{sec:verify}

\begin{table}[h]
\centering
\small
\begin{tabular}{lcccc}
\toprule
& \multicolumn{2}{c}{\tblite{}} & \multicolumn{2}{c}{\tbtwo{}} \\
\cmidrule(lr){2-3}\cmidrule(lr){4-5}
Verifier & Best-of-32 & Gap recovered & Best-of-32 & Gap recovered \\
\midrule
Random (pass@1)  & 42.0 & 0\%   & 21.6 & 0\%   \\
\qwennine{}      & 54.2 & 28\%  & 35.8 & 41\%  \\
\tmaxnine{}      & 55.4 & 31\%  & 37.2 & 45\%  \\
\glm{}           & 64.6 & 53\%  & 44.0 & 65\%  \\
Oracle (pass@32) & 85.0 & 100\% & 56.2 & 100\% \\
\bottomrule
\end{tabular}
\caption{\textbf{Verifier recoverability at $k{=}32$.} Candidate pools generated by \qwennine{}. Gap recovered $=$ (best-of-32 $-$ pass@1) / (pass@32 $-$ pass@1). Figure~\ref{fig:verification} shows the full curves over $k$.}
\label{tab:verification}
\end{table}

\begin{figure*}[h]
    \centering
    \includegraphics[width=0.8\linewidth]{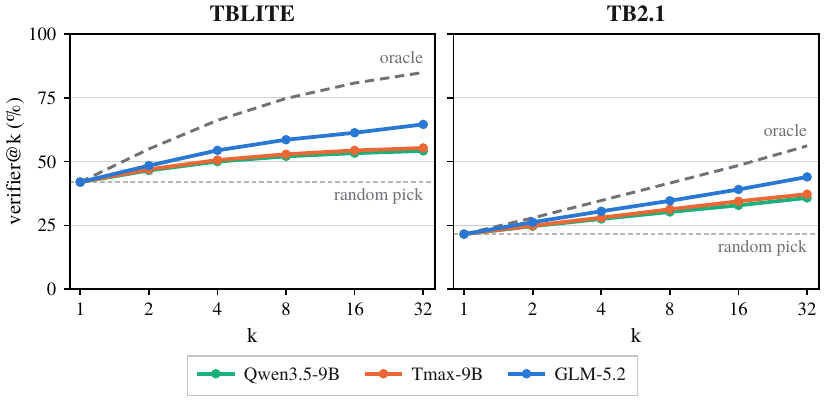}
    \caption{\textbf{SLMs lag behind \glm{} in outcome verification.} Given the same pool of candidate trajectories, SLMs recover less of the random--oracle gap. On \tblite{}, most gains occur by $k{\approx}8$; on \tbtwo{}, gains continue and the gap widens as $k$ grows.}
    \label{fig:verification}
\end{figure*}

Solvability shows that successful trajectories exist in the sample pool;
whether a model can \emph{find} them is a distinctly different capability. To measure it,
we evaluate each model as an outcome verifier: for each task we sample $n{=}32$
candidate trajectories from \qwennine{}, then prompt each model as a
verifier to select the candidate most likely to have passed the task's grader. Each trajectory is scored pointwise and then the highest-scoring trajectory is selected
(best-of-$k$). \emph{Verifier@$k$} is the pass rate of the candidate selected from $k$ candidates, averaged over all size-$k$ subsets of the pool. Verifier@$k$ is bounded below by random selection, that is, the pool's mean
pass rate (pass@1), and above by oracle selection (pass@$k$). Figure~\ref{fig:verification} shows that every model's curve lies in this range. We provide additional implementation details in Appendix~\ref{app:verification}.

\textbf{Results.} At $k{=}32$, \glm{} recovers 53\% of the random--oracle gap on \tblite{}, compared with 28--31\% for the SLMs (Table~\ref{tab:verification}). The same ordering holds on \tbtwo{}, where \glm{} recovers 65\% of the gap, compared with 41--45\% for the SLMs. On \tblite{}, most gains occur by $k{\approx}8$ (Figure~\ref{fig:verification}); on \tbtwo{}, gains continue through $k{=}32$ and the absolute gap between \glm{} and \tmaxnine{} widens from 3.3 points at $k{=}8$ to 6.8 at $k{=}32$. On pools generated by \tmaxnine{} and \glm{}, across both benchmarks, SLM verifiers recover 22--47\% of the
random--oracle gap, while \glm{} recovers 52--74\%
(Appendix~\ref{app:verification}, Table~\ref{tab:other-pools}). Thus, while repeated sampling narrows the \mbox{SLM--FM} gap in solvability (\S\ref{sec:passk}), a persistent gap remains in verification: \textbf{small models can generate successful solutions to terminal tasks, yet remain substantially worse at identifying them among candidate trajectories.}


\subsection{Model Confidence}
\label{sec:confidence}

A capable agent must be able to estimate its probability of success in
order to allocate resources such as the number of turns, thinking effort, or escalation to a stronger model. We therefore analyze the
\mbox{SLM--FM} gap in this capability, eliciting prospective confidence
\citep{barkan2025largelanguagemodelsknow}. Each model estimates its probability of solving a task given only the task description before taking any action. We provide additional implementation details in Appendix~\ref{app:confidence}.

\begin{wrapfigure}{R}{0.5\textwidth}
    \centering
    \includegraphics[width=\linewidth]{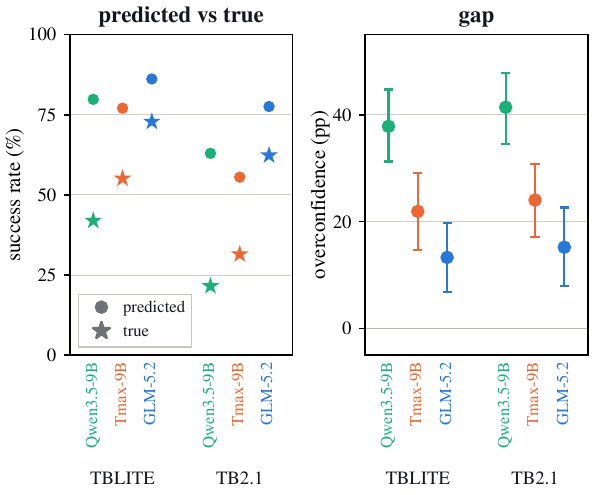}
    \caption{\textbf{Prospective confidence.} Left: mean predicted vs.\ true success rate per model and benchmark. Right: overconfidence, the mean predicted minus true success rate; error bars show 95\% bootstrap intervals (Appendix~\ref{app:confidence}).}
    \label{fig:confidence}
\end{wrapfigure}

\textbf{Results.} 
Figure~\ref{fig:confidence} reports two views of
these confidence estimates: mean predicted versus true success
rate per model and overconfidence, the mean predicted
minus true success rate. We find that all models are overconfident on both benchmarks, but the degree of
miscalibration varies across models: \glm{} is the least overconfident and \qwennine{} the most. On \tblite{}, the gap
between predicted and true success is 13 percentage points for \glm{},
compared with 22 points for \tmaxnine{} and 38 points for \qwennine{}.
On \tbtwo{}, the corresponding gaps are 15, 24, and 41 points; on this harder benchmark, all models lower their predicted success, but by less than their true success drops, so overconfidence grows by 2--3 points. Consistent
with these differences, \glm{} achieves a substantially lower Brier score
(0.21) than the SLMs (0.30--0.40) on \tblite{} (Figure~\ref{fig:summary}).
\textbf{Thus, SLMs are less calibrated in their estimates of their own probability of success, making these estimates unreliable for test-time allocation decisions.}

\section{Discussion and Future Work}
We investigate why small language models trail frontier models as terminal agents. Rather than treating task success as a single capability, we decompose the \mbox{SLM--FM} gap into solvability, verification, and confidence. Across two benchmarks, we find that these axes reveal complementary aspects of the gap. Repeated sampling substantially narrows the gap in solvability, showing that some SLM failures reflect unreliable generation rather than an inability to reach a correct solution. However, the gap is not eliminated, particularly on the harder \tbtwo{} benchmark. Beyond generation, SLMs remain substantially worse at selecting successful trajectories from a shared candidate pool and at estimating their probability of success before acting.
\textbf{Our results show that the \mbox{SLM--FM} gap is multidimensional: closing it requires improving not only the ability to solve terminal tasks, but also the ability to recognize successful solutions and anticipate success.}

Our findings raise several directions for future work. A natural next step is to study how solvability, verification, and confidence interact in an agentic loop: for example, whether calibrated uncertainty can help an SLM decide when to sample again, verify a trajectory, or escalate to a stronger model. It is also important to extend the comparison to additional frontier models and to evaluate verification on more diverse candidate pools beyond the generator models in Table~\ref{tab:other-pools} (Appendix~\ref{app:verification}), including trajectories from other harnesses. A further direction is execution-based verification, in which the verifier writes and runs
its own tests against the final environment state rather than judging the trajectory from
text alone; this would separate a model's ability to verify from the limits of
prompted, text-only judgment. Finally, developing training methods that directly improve verification and confidence, rather than relying on these capabilities to emerge from improvements in generation, remains an open direction.

\bibliography{references}
\bibliographystyle{plainnat}


\clearpage
\appendix
\section{LLM Usage Disclosure}
Claude and ChatGPT were used to write code for the experiments and to edit the manuscript. The project idea, experimental design, and writing are primarily the authors' own work.

\section{Benchmark Difficulty}
\label{app:difficulty}
\begin{figure}[h]
    \centering
    \includegraphics[width=\linewidth]{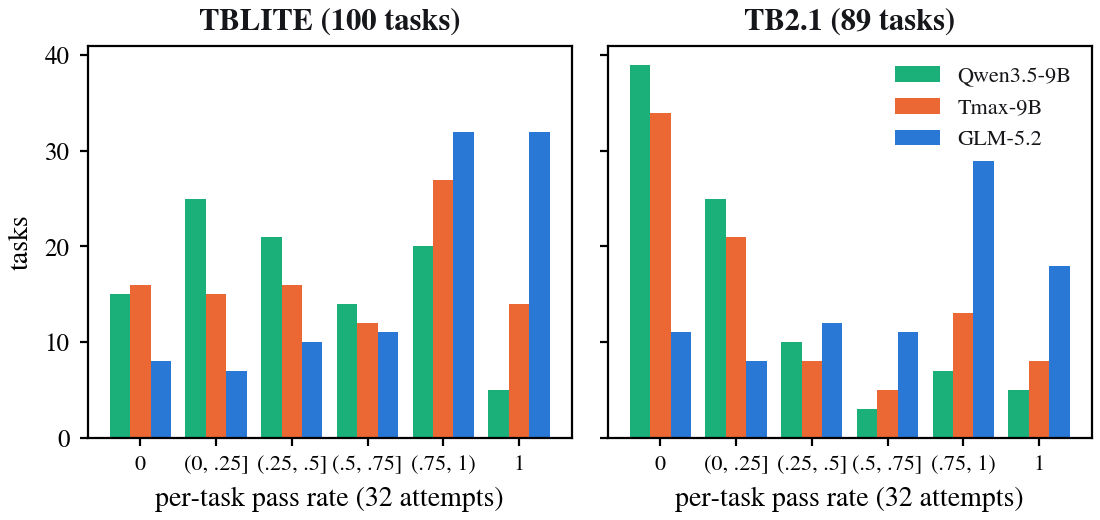}
    \caption{\textbf{Distribution of per-task pass rates over up to 32 attempts for each model on \tblite{} and \tbtwo{}.}}
    \label{fig:difficulty}
\end{figure}

Figure~\ref{fig:difficulty} shows, for each model, the distribution of per-task pass rates over up to 32 attempts on each benchmark. On \tbtwo{}, many more tasks are never solved by the SLMs than by \glm{}, consistent with \tbtwo{} being harder, particularly for SLMs (\S\ref{sec:prelim}).

\section{Harness Choice Changes Model Performance}
\label{sec:harness-reward}
Table~\ref{tab:models} lists the models we evaluate. We evaluate each model with four popular agent harnesses and describe them below:

\begin{table}[h]
\centering
\small
\setlength{\tabcolsep}{5pt}
\begin{tabular}{@{}lllll@{}}
\toprule
Model & Parameters & Architecture & Checkpoint & Tier \\
\midrule
Qwen3.5-9B & 9B                & dense        & \texttt{Qwen/Qwen3.5-9B}     & SLM \\
Tmax-9B    & 9B                & dense        & \texttt{allenai/tmax-9b}     & SLM \\
GLM-5.2    & 744B (40B active) & MoE, 1M ctx  & \texttt{zai-org/GLM-5.2-FP8} & FM \\
\bottomrule
\end{tabular}
\caption{\textbf{Models.} SLMs and the FM used in this study.}
\label{tab:models}
\end{table}

\begin{itemize}
    \item \textbf{\minisa{}}~\citep{yang2024sweagent}: A minimal agent that Harbor installs and runs as a command-line tool. It does not include compaction.
    \item \textbf{\vanillux{}}~\citep{ivison2026tmax}: A harness optimized for SLMs that builds on mini-swe-agent prompts. It expects native bash tool calls, parsed by vLLM and executed via Docker. The shell state is replayed via a persistent shim at each turn. It does not include compaction and terminates on timeout errors. It is built to be robust to formatting errors.
    \item \textbf{\terminus{}}~\citep{merrill2026terminal}: Harbor's built-in terminal agent. It expects JSON blobs as free text, parsed with formatting auto-repair. Commands are executed with a tmux live pane and a fully persistent login shell. Command observations are captured as screen-diffs between turns. When fewer than 8k tokens of context remain, it compacts the history through a summarization handoff that uses three LLM calls.
    \item \textbf{\openhands{}}~\citep{wang2025openhandsopenplatformai}: Third-party SDK with native tool calling: terminal, file\_editor, task\_tracker (+ MCP, skills). Commands are run in persistent terminal shells and observations are shown as structured tool outputs to the model. Compaction is implemented with a native condenser supplied with the SDK.\footnote{We upgraded 49 task images to Python 3.12 to run openhands-sdk; this did not affect task outcomes or definitions.}
\end{itemize}

Table~\ref{tab:harness_effects} reports mean reward per model-harness pair across 3 trials with a context length of 256k.

\textbf{No single harness is best for all models.} Harness choice affects model performance: \qwennine{} drops from 43.1 under \vanillux{} to 25.0 under \openhands{} in Table~\ref{tab:harness_effects}. Even frontier models like \glm{} show variable performance under each harness. The best harness also differs per model: \glm{} performs best on \terminus{}, while \tmaxnine{} performs worst on \terminus{}. This is likely due to the brittle nature of \terminus{} that requires small models to send keystrokes to a live terminal and the complex tool formats that are harder for smaller models to follow.

\begin{table}[h]
\centering
\begin{tabular}{lrrrr}
\toprule
 & \minisa{} & \terminus{} & \vanillux{} & \openhands{}\\
\midrule
\qwennine{}  & $43.0 \pm 5.1$ & $40.1 \pm 0.1$ & $\textbf{43.1} \pm 0.7$ & $25.0 \pm 1.9$\\
\tmaxnine{}      & $\textbf{57.7} \pm 1.8$ & $47.1 \pm 3.9$ & $55.1 \pm 3.2$ & $49.5 \pm 1.5$\\
\glm{}      & $75.8 \pm 0.7$ & $\textbf{84.2} \pm 5.1$ & $79.7 \pm 4.1$ & $72.8 \pm 2.1$\\
\midrule
Average      & $58.8$ & $57.1$ & $\textbf{59.3}$ & $49.1$\\
\bottomrule
\end{tabular}
\caption{\textbf{Mean reward (\%) on \tblite{} across harnesses, averaged over 3 trials}. All models are served at 256k context length.
}

\label{tab:harness_effects}
\end{table}

\section{Verifier Setup}
\label{app:verification}
Each candidate trajectory is serialized as the task definition followed by the visible message, command, and output from every agent step; the agent's private reasoning is excluded. Inputs exceeding the token budget (48k on \tblite{} and 128k on \tbtwo{}) are truncated by retaining the first 25\% and last 75\% of the budget. The verifier is prompted to write a brief analysis and assign a final pointwise score between 0 and 1 (see the verifier prompt below). Candidates that remain unscored due to parsing failures are ranked below all scored candidates. Finally, the highest-scoring candidate is selected, with ties broken uniformly at random. We compute verifier@$k$ exactly by taking the expectation over all $\binom{32}{k}$ candidate subsets for each task.

\begin{tcolorbox}[title=Verifier prompt, colback=gray!5, colframe=gray!60,
  fonttitle=\small\bfseries, left=2mm, right=2mm, top=1mm, bottom=1mm]
\footnotesize
\textbf{System:}
\begin{verbatim}
You are a strict verifier of terminal-agent work.
\end{verbatim}
\textbf{User:}
\begin{verbatim}
Below is a terminal task and the full log of an agent's attempt (commands and
outputs; the agent's private reasoning is not shown). Decide whether the
attempt actually completed the task as specified.

Rules:
- Judge only from evidence in the log: files written, commands run, outputs.
- Do not trust the agent's own claims of success; verify them against the log.
- Consider every requirement in the task, including output formats and file paths.

First write a brief analysis (at most 300 words) of what the task requires and
what the log shows. Then output one final line exactly of the form:
SCORE: <number between 0.00 and 1.00>
where 1.00 means the task is certainly completed and 0.00 means it certainly
is not. Use intermediate values for partial or uncertain completion.

{trajectory}
\end{verbatim}
\end{tcolorbox}

\textbf{Other candidate pools.} Table~\ref{tab:other-pools} repeats the verification experiment of \S\ref{sec:verify} on candidate pools generated by \tmaxnine{} and \glm{}. On every pool and both benchmarks, \glm{} recovers more of the random--oracle gap than either SLM verifier.

\begin{table}[h]
\centering
\small
\begin{tabular}{llccc}
\toprule
& & \multicolumn{3}{c}{Verifier} \\
\cmidrule(lr){3-5}
Benchmark & Pool generated by & \qwennine{} & \tmaxnine{} & \glm{} \\
\midrule
\multirow{3}{*}{\tblite{}}
 & \qwennine{} & 28\% & 31\% & 53\% \\
 & \tmaxnine{} & 22\% & 23\% & 72\% \\
 & \glm{}      & 31\% & 32\%  & 74\% \\
\midrule
\multirow{3}{*}{\tbtwo{}}
 & \qwennine{} & 41\% & 45\% & 65\% \\
 & \tmaxnine{} & 41\% & 44\% & 52\%  \\
 & \glm{}      & 44\% & 47\%  & 52\% \\
\bottomrule
\end{tabular}
\caption{\textbf{Verification across candidate pools.} Share of the random--oracle
gap recovered at $k{=}32$ by each verifier (columns) on pools generated by each model.}
\label{tab:other-pools}
\end{table}

\section{Confidence Elicitation and Calibration}
\label{app:confidence}
Before attempting each task, we elicit each model's predicted probability of successfully completing it (see the confidence prompt below). The model receives only the task instruction, wrapped in a prompt asking it to estimate its probability of success and return a single number between 0 and 1. No environment access is provided during elicitation, and no system prompt is used. We collect five estimates per task at temperature 1.0 and use their mean as the task-level confidence.

\textbf{True success rate.} We estimate each model's true success rate from its own repeated-sampling attempts on the same tasks (\S\ref{sec:passk}). Errored trials count as failures. We compute the true success rate by averaging per-task success rates, giving each task equal weight.

\textbf{Calibration metrics.} We measure calibration using the Brier score and overconfidence. The Brier score is computed over task–trial pairs as the mean squared difference between the task-level predicted confidence and each trial's binary outcome. Overconfidence is the mean difference between each task's predicted confidence and its empirical success rate, averaged equally across tasks. 

\textbf{Uncertainty estimation.} We report 95\% percentile bootstrap intervals based on 1,000 task-level bootstrap resamples with replacement (random seed 7). This procedure preserves task-level clustering and is used to estimate uncertainty in the reported metrics.

\begin{tcolorbox}[title=Confidence prompt, colback=gray!5, colframe=gray!60,
  fonttitle=\small\bfseries, left=2mm, right=2mm, top=1mm, bottom=1mm]
\footnotesize
\begin{verbatim}
Here is a task you will be asked to complete in a terminal environment.

{instruction}

Before starting, estimate the probability that you will complete this task
successfully. Reply with a single number between 0 and 1 and nothing else.

Output format: 0.XX
\end{verbatim}
\end{tcolorbox}

\section{Additional Failure Analysis}
\label{app:failure}
\begin{figure}[p]
    \centering
    \includegraphics[width=\linewidth]{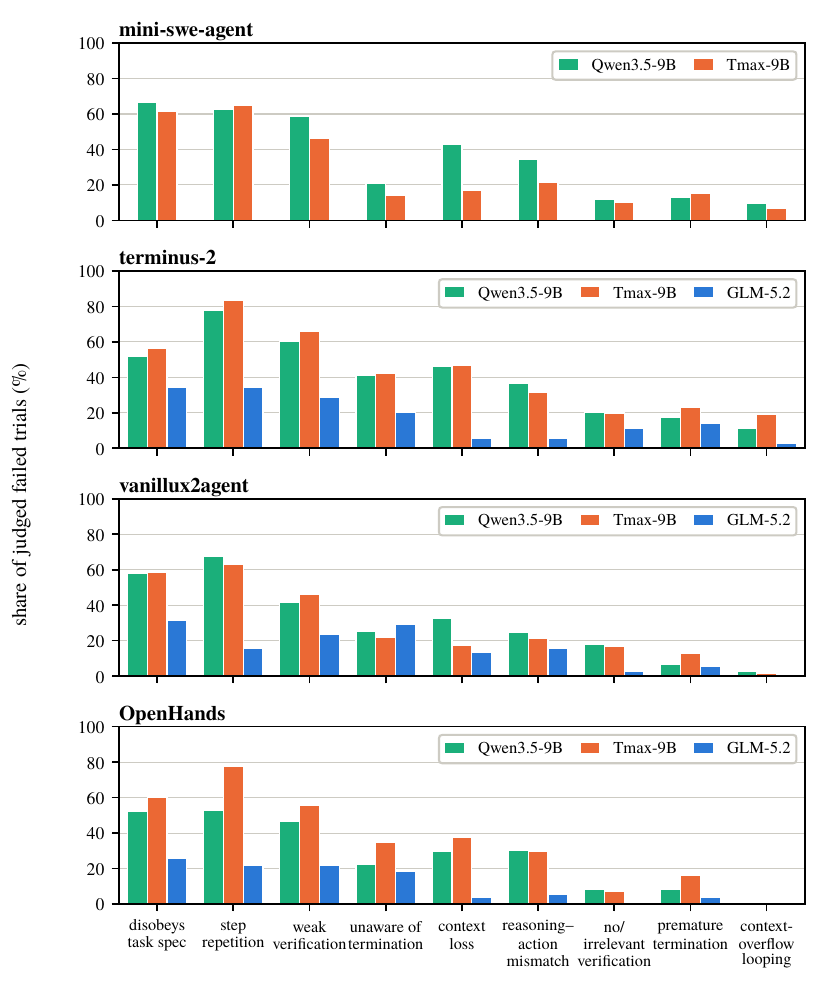}
    \caption{\textbf{Failure modes across various harnesses.} Prevalence of failure modes across four widely used harnesses is shown on \tblite{}. GLM-5.2 failures were not judged under \minisa{}.}
    \label{fig:fail_harness}
\end{figure}

We classify the failed trials into the failure modes of Table~\ref{tab:fail_modes}, following the taxonomy of \citet{merrill2026terminal}, with an LLM judge (GPT-5, high reasoning). For each model and harness, we judge a sample of its failed trials, dropping trials with no computed reward due to infrastructure errors. Figure~\ref{fig:fail_harness} shows the failure modes across all harnesses, including our SLM-specific mode 09 (context-overflow looping), which Figure~\ref{fig:failure_vanillux} omits; it is rare under \vanillux{} (at most 2.8\% of failed trials).

\begin{table}[t]
\centering
\small
\begin{tabular}{@{}llp{9.0cm}@{}}
\toprule
ID & Failure mode & Definition (judged MATCH/NO\_MATCH per failed trajectory) \\
\midrule
01 & Disobey Task Specification & The trajectory materially contradicts an explicit directive of the task (a hard requirement, prohibition, or binding default). \\
02 & Step Repetition & The agent re-executes an already-attempted phase (same sub-goal, tool, and target) with semantically identical actions, absent any material change in state. \\
03 & Unaware of Termination & The agent continues acting beyond a reasonable stopping point—after success is achieved or futility is established. \\
04 & Context Loss & The agent forgets or contradicts relevant recent context: concrete state (files created, errors seen) or its own earlier decisions. \\
05 & Premature Termination & The agent declares completion before meeting explicit task objectives or supplying required verifications or critical information. \\
06 & No/Irrelevant Verification & The agent marks the task complete with no check of core functional properties, or with only irrelevant checks. \\
07 & Reasoning--Action Mismatch & The agent's stated reasoning, claims, or commitments conflict with its observable actions and produced artifacts. \\
08 & Weak Verification & Verification is present but does not cover the properties that must hold for true correctness (includes fabricated evidence of success). \\
09 & Context-Overflow Looping & After a context-limit event, the agent proposes a ``new'' approach that is logically identical to a previously failed one (SLM extension; not in TB2). \\
\bottomrule
\end{tabular}
\caption{\textbf{Failure-mode taxonomy.} Modes 01--08 adapted from the Terminal-Bench~2
methodology~\cite{merrill2026terminal}; mode 09 is our SLM-specific extension. A single frontier judge
(GPT-5, high reasoning effort) scores each mode independently on failed
trajectories, so modes are non-exclusive.}
\label{tab:fail_modes}
\end{table}
\FloatBarrier

\end{document}